\pdfoutput=1

\documentclass[11pt]{article}

\usepackage[]{EMNLP2022}

\usepackage{times}
\usepackage{latexsym}

\usepackage[T1]{fontenc}

\usepackage[utf8]{inputenc}

\usepackage{microtype}

\usepackage{inconsolata}

\usepackage[ruled]{algorithm2e}
\usepackage{booktabs}
\usepackage{graphicx}
\usepackage{subfigure}
\usepackage{multirow}
\usepackage{multicol}
\usepackage{hyperref}
\usepackage{cite}
\usepackage{makecell}
\usepackage[flushleft]{threeparttable}
\usepackage{amsmath,amssymb,amsfonts}
\usepackage{scalerel,xparse}
\usepackage{algorithmic}

\newcommand{\supportset}{$D_{support}$}
\newcommand{\randomtotally}{\textbf{TR}}
\newcommand{\randomsupport}{\textbf{SR}}
\newcommand{\standardnol}{\textbf{SN}}
\newcommand{\standardwrong}{\textbf{SW}}
\newcommand{\standard}{\textbf{ST}}

\title{Robustness of Demonstration-based Learning\\ Under Limited Data Scenario}

\author{Hongxin Zhang$^1$, Yanzhe Zhang$^2$, Ruiyi Zhang$^3$, Diyi Yang$^4$ \\
   $^1$Shanghai Jiao Tong University, $^2$Georgia Institute of Technology \\ $^3$Adobe Research, $^4$Stanford University  \\
  \texttt{$^1$icefox@sjtu.edu.cn, $^2$z\_yanzhe@gatech.edu}\\ \texttt{$^3$ruizhang@adobe.com, $^4$diyiy@cs.stanford.edu }}

\begin{document}
\maketitle

\begin{abstract}
Demonstration-based learning has shown great potential in stimulating pretrained language models’ ability under limited data scenario.
Simply augmenting the input with some demonstrations can significantly improve performance  
on few-shot NER. 
However, why such demonstrations are beneficial remains unclear since there is no explicit alignment between the demonstrations and the predictions. 
In this paper, we design pathological demonstrations by gradually removing intuitively useful information from the standard ones to take a deep dive of the robustness of demonstration-based sequence labeling and show that
(1) 
demonstrations composed of random tokens still make the model a better few-shot learner;  
(2) the length of random demonstrations and the relevance of random tokens are the main factors affecting the performance;
(3) demonstrations increase the confidence of model predictions on captured superficial patterns.
We have publicly released our code at \url{https://github.com/SALT-NLP/RobustDemo}.
\end{abstract}
\section{Introduction}
Current large pretrained language models (PLMs) struggle to learn NLP tasks under limited data scenarios \citep{devlin-etal-2019-bert, DBLP:journals/corr/abs-1907-11692, lewis-etal-2020-bart,XieSemi,huang-etal-2021-shot}. In contrast,  humans can solve natural language tasks with only a few illustrative examples\citep{lake2015human}.
Motivated by this, demonstration-based learning has been introduced to augment the input with a few examples and labels.
For instance, \citet{NEURIPS2020_1457c0d6} simply picked up to 32 randomly sampled instances and directly concatenated them with the input to perform \emph{in-context learning} with the model frozen and significantly boosted the performance. \citet{lee-etal-2022-good} concatenated the input with task demonstrations to create augmented input and fed them into PLMs to obtain improved token representations to do sequence labeling in a classifier-based fine-tuning way.

\begin{figure}[t]
    \definecolor{camel}{rgb}{0.76, 0.6, 0.42}
    \centering
    \resizebox{1.0\columnwidth}{!}{\includegraphics{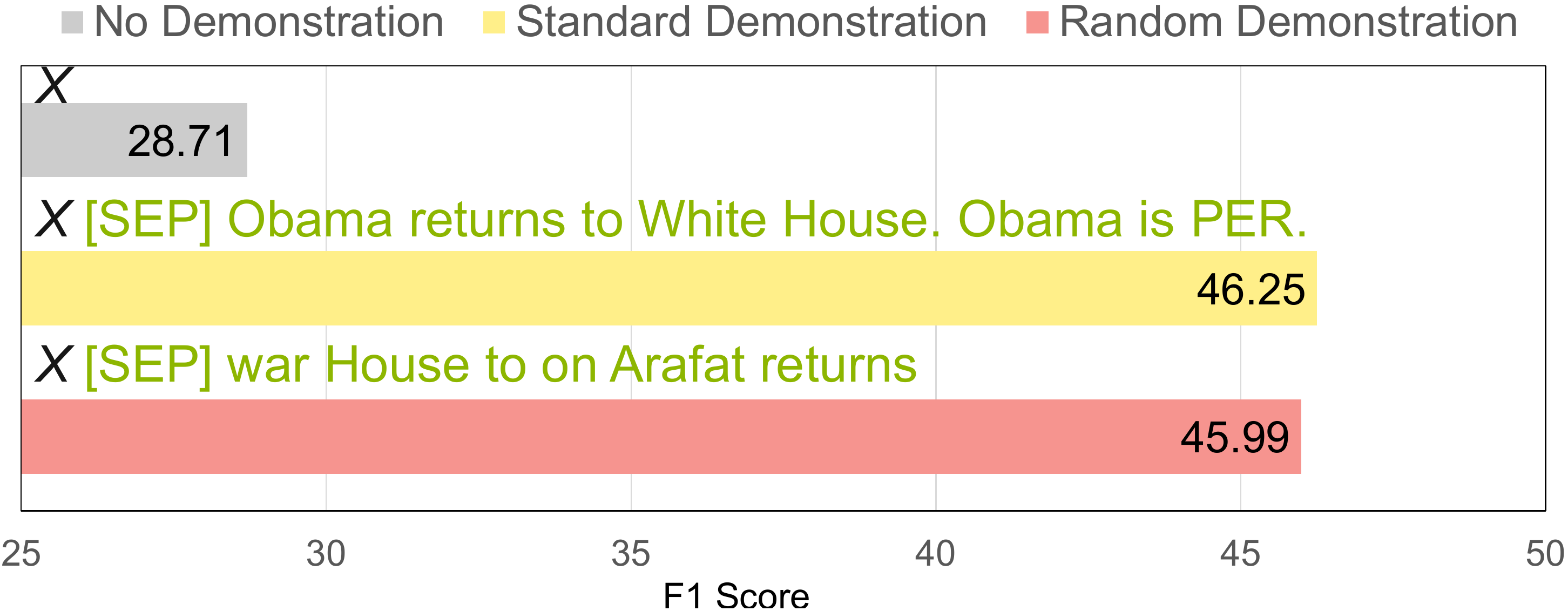}}
    \caption{\textbf{Performance for different demonstrations on CoNLL03 5-shot support set.} Here \textit{X} denotes the input sentence, such as "Jobs was born in America.", and we show the PER part of the whole demonstration in {\definecolor{applegreen}{rgb}{0.55, 0.71, 0.0}\color{applegreen} applegreen} for visualization. 
    Surprisingly, random tokens can be good demonstrations too.}
    \label{fig:intro}
\end{figure}

However, how and why such demonstrations help remains unclear. As such, there has been a growing amount of work investigating the robustness and interpretability of demonstration-based learning. 
For instance, \citet{lu2021fantastically} reported that few-shot text classification is very sensitive to the ordering of the demonstrations in in-context learning. On a wide range of low-resource Natural Language Understanding (NLU) tasks, \citet{min2022rethinking} investigated why demonstrations in in-context learning can bring performance gains over zero-shot inference and found that correct input-label mapping matters very little. 

Building on these prior works, we take a deeper dive into the robustness of demonstration-based learning \citep{lee-etal-2022-good}, especially for structured prediction tasks like Named Entity Recognition (NER).
Demonstrations might not be robust for more structured prediction settings since these limited amounts of examples might not include much inductive bias.
Also, using classifier-based fine-tuning demonstrations could be even more unreliable since there is no alignment between the demonstrations and the prediction space.



Concretely, we investigate the robustness of demonstration-based sequence labeling by designing pathological demonstrations: gradually ruling out the helpful information from the demonstrations. We surprisingly find that a working demonstration does not need to contain correct examples to observe improved performance.
Furthermore, randomly replacing every token in the demonstration can still make a better few-shot learner even it is no longer a meaningful sentence and does not make any sense (Figure \ref{fig:intro}).
This observation conflicts with some existing hypotheses \citep{gao-etal-2021-making, lee-etal-2022-good} that models are learning meaningful knowledge from these demonstrations.
We also find that the length of the pathological demonstration and the relevance of its random tokens drastically affect the performance.
Empirical results on Name Regularity Bias (NRB) diagnose dataset~\citep{NRB} shows that the demonstrations rarely help the performance when there is no easy patterns. 
Additionally, we show the pathological demonstrations can obtain similar or better performance on NLU tasks such as classification and natural language inference.
In summary, our empirical results encourage the rethinking on how the demonstration helps the model obtain better few-shot capability and provides some insights.
\begin{figure*}[t]
    \centering
    \subfigure[Traditional Token Classification]{
        \includegraphics[width=0.45\textwidth]{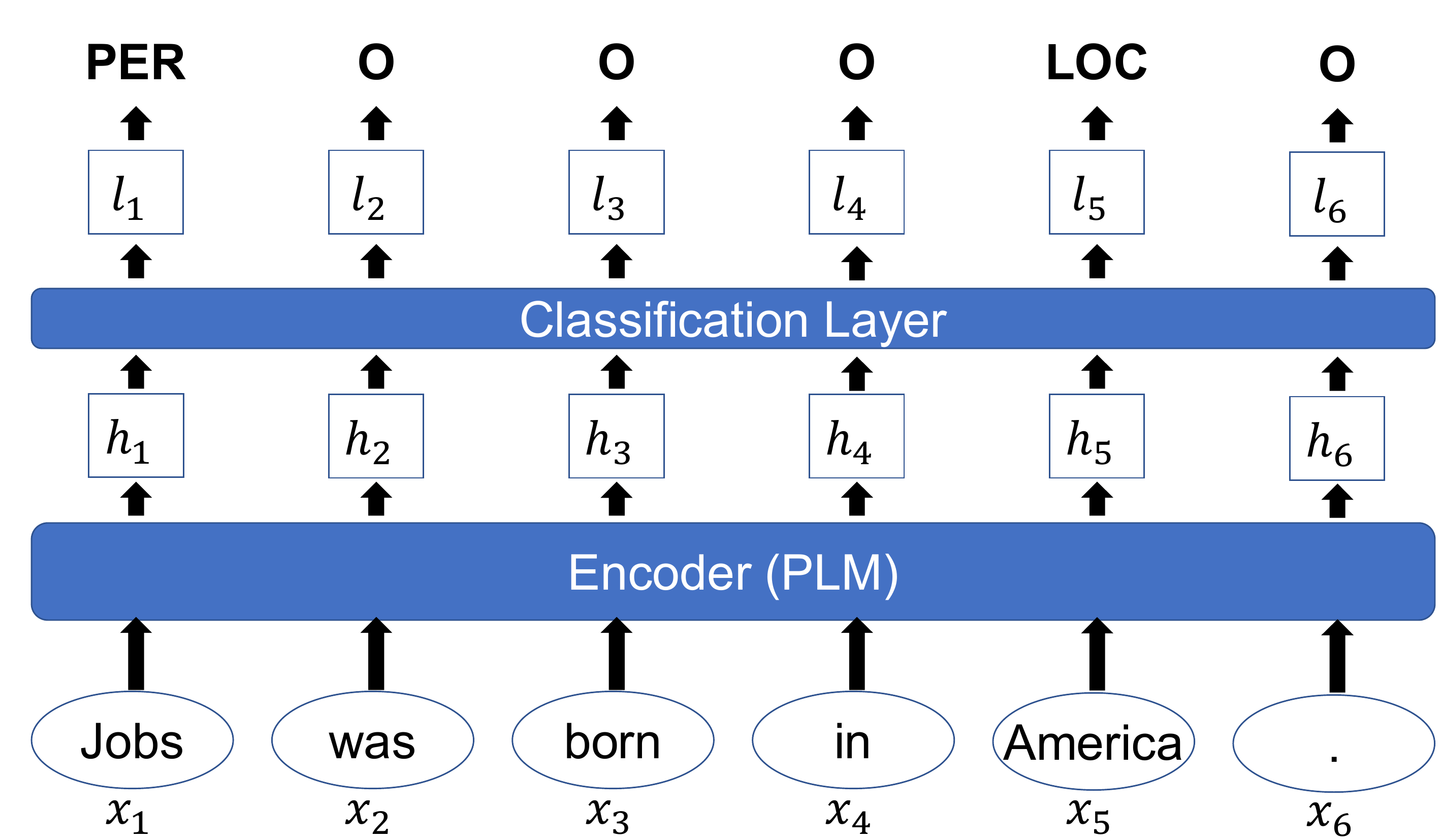}
        \label{fig:traditional}
    }
    \subfigure[Task Demonstration Construction]{
        \includegraphics[width=0.5\textwidth]{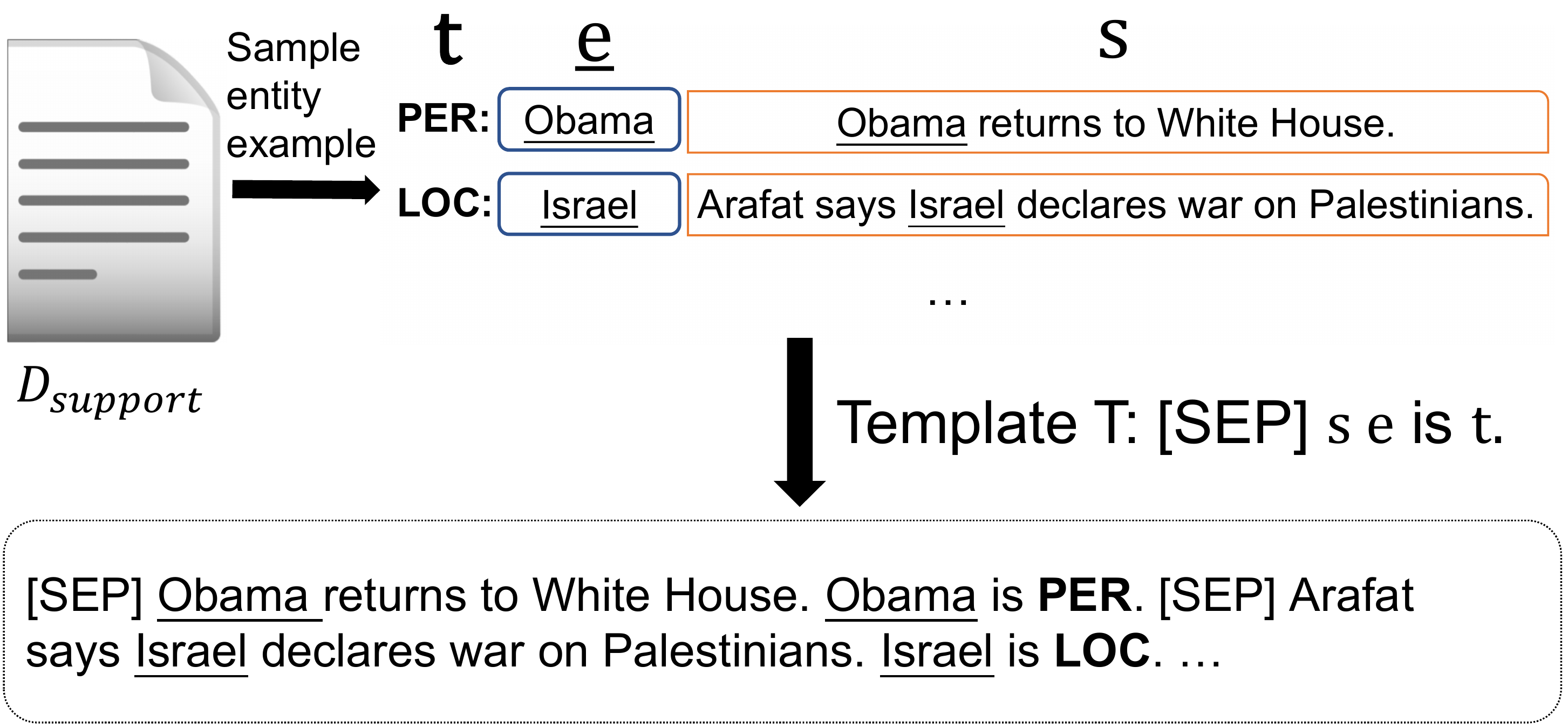}
        \label{fig:construction}
    }
    \subfigure[Demonstration-based Learning]{
        \includegraphics[width=0.95\textwidth]{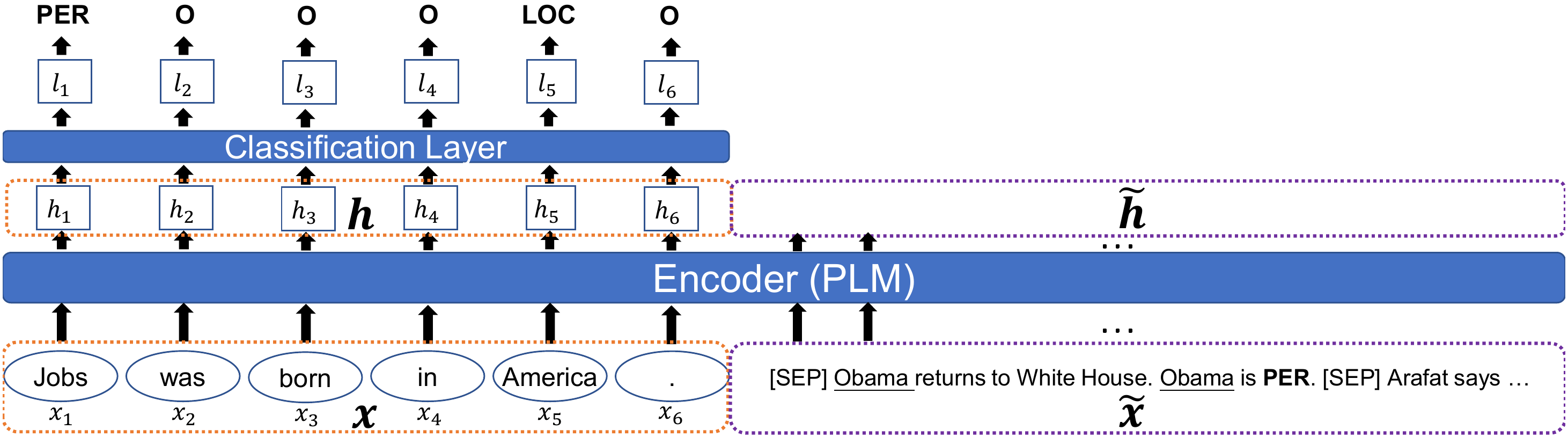}
        \label{fig:demonstration-based learning}
    }
    \caption{An overview of (a) traditional token classification methods, (b) task demonstration construction process, and (c) demonstration-based learning for NER.}
    \label{fig:overview}
\end{figure*}

\section{Related Work}

\subsection{Demonstration-based Learning}
Demonstrations are first introduced by the GPT series \citep{radford2019language,NEURIPS2020_1457c0d6}, where a few examples are sampled from training data and transformed with templates into appropriately-filled prompts.  The existing demonstration-based learning research can be broadly divided into three categories based on the task reformulation and whether there is an update on the model parameters:

\paragraph{In-context Learning} reformulates the task as a language modeling problem, and the model makes predictions by filling in the blank without using classifiers. In in-context learning, the model learns by only conditioning on the demonstration in a tuning-free way, while an enormous language model is often needed for this method \citep{NEURIPS2020_1457c0d6,zhao2021calibrate,min-etal-2022-noisy, wei2022chain}.

\paragraph{Prompt-based Fine-tuning} also reformulates the task as a masked language modeling problem, and the demonstrations are incorporated as additional contexts \citep{gao-etal-2021-making}. The model learns by fine-tuning on a small set of training data and moderately-sized PLMs are often used. Virtual demonstrations such as trainable embeddings \citep{liang2022contrastive} belong to this setting.

\paragraph{Classifier-based Fine-tuning} requires no reformulation of the task and simply augments the original input with demonstrations \citep{lee-etal-2022-good}. One advantage of this method is that it can benefit tasks such as sequence labeling when it is hard to be reformulated as (masked) language modeling.

Much work has been conducted to examine how to make a good sample selection \citep{liu2021makes, mishra2021cross} and ordering \citep{lu2021fantastically} as informative demonstrations are crucial for model performance.
Our work focuses on the third demonstration-based learning method---classifier-based fine-tuning under the traditional token classification framework on sequence labeling tasks. 

\subsection{Analyses of Prompts and Demonstrations}
With the recent prevalence of using prompts and demonstrations to stimulate the ability of PLMs under limited data scenarios \citep{schick-schutze-2021-exploiting, schick-schutze-2021-just,liu2021pre}, a growing amount of works look at how prompting and demonstrations work. For instance, \citet{webson2021prompt} studied different prompt templates and target words on NLI tasks mainly with the prompt-based fine-tuning method and found evidence that prompt-based models still perform well when irrelevant or even misleading prompts are used. Similarly, \citet{min2022rethinking} showed that in-context learning is not taking advantage of correct label mappings but more surface form in the demonstrations, like the distribution of the input text and the overall format of the sequence, while later work \citep{kim2022groundtruth} argued the impact of correct mapping depends on different configurations.
\citet{logan2021cutting} demonstrated fine-tuning language models on a few data can considerably reduce the need for prompt engineering, while \citet{utama-etal-2021-avoiding} showed that prompt-based fine-tuning improves the in-distribution performance while gradually increases models' reliance on surface heuristics.
\citet{garg2022transformers} trained transformers to directly learn function classes provided in demonstrations and showed such learning is possible even under distribution shifts.

Different from them, we focus on the analysis of demonstration-based learning under the classifier-based fine-tuning framework on sequence labeling tasks where we do not reformulate the task into (masked) language model problems and therefore rule out the impact of target word selection. Furthermore, we create more effective and adversarial demonstrations consisting of only random tokens, showing that it is not only the in-context learning or prompt-based fine-tuning but also the traditional ways of utilizing PLMs that may trigger this counter-intuitive performance gain.
\section{Problem Definition}
This section focuses on demonstration-based learning under limited data scenario,  by introducing concepts of limited data sequence labeling tasks in Section~\ref{FSNER}, as well as describing traditional token classification methods in Section~\ref{TTC} and  demonstration-based learning in Section~\ref{DemoNER}.

\subsection{Limited Data Sequence Labeling Tasks}
\label{FSNER}
Given an input sentence $\mathbf{x} = [x_1,x_2,\cdots,x_n]$ composed of $n$ tokens, the sequence labeling task is to predict a tag $y_i\in Y \cup \{O\}$ for each token $x_i$, where $Y$ is a predefined set of tags such as \{LOC, PER, ...\} for Named Entity Recognition (NER) and \{NP,  VP, ...\} for chunking, and $O$ denotes outside a tagged span. Under limited data scenario, we only have $K$-shot support set \supportset\ for training which contains $K$ examples for each tag type.

\subsection{Traditional Token Classification Methods}
\label{TTC}

As shown in Figure~\ref{fig:traditional}, traditional methods for sequence labeling use the encoders from PLMs such as BERT to encode the input $\mathbf{x} = [x_1,x_2,\cdots,x_n]$ to get contextualized representations for each token $\mathbf{h} = [h_1,h_2,\cdots,h_n]$, and then use a linear classifier or CRF layer to get the label estimation $l_i$ for each token. The model is trained to minimize the cross entropy loss between $l_i$ and $y_i$.

\subsection{Demonstration-based Learning}
\label{DemoNER}
\paragraph{Constructing Task Demonstration}
As shown in Figure~\ref{fig:construction}, for each tag type $t^{(c)}$, we sample one tag example $e^{(c)}$, along with its original context $s^{(c)}$ from support set \supportset\ . We use template $T$ to convert  them into type demonstration $d^{(c)} = T(s^{(c)}, e^{(c)}, t^{(c)})$, and then construct task demonstration $\tilde{\mathbf{x}}$ by concatenating all type demonstrations together:
$$\tilde{\mathbf{x}} = d^{(1)} \oplus d^{(2)} \oplus \cdots \oplus d^{(|Y|)}$$
Here $\oplus$ denotes the concatenation of input sequences. We further concatenate the original input $\mathbf{x}$ with demonstration $\tilde{\mathbf{x}}$ to obtain demonstration-augmented input $[\mathbf{x};\tilde{\mathbf{x}}]$.
Prior work 
\citep{lee-etal-2022-good} have studied various example sampling strategies and templates to construct the demonstration. 
We adopt their \texttt{popular} strategy to choose $e^{(c)}$ that occurs most frequently among the corresponding examples and \texttt{context} template of "$s^{(c)}$. $e^{(c)}$ is $t^{(c)}$.", given their strong performances in \citet{lee-etal-2022-good}. Here, we refer the "$e^{(c)}$ is $t^{(c)}$." part in the template as labeling part of the demonstration.

\paragraph{Learning with Demonstration}
As shown in Figure~\ref{fig:demonstration-based learning}, like traditional token classification methods, we feed the demonstration-augmented input $[\mathbf{x};\tilde{\mathbf{x}}]$ into the encoder, and get the token representation $[\mathbf{h};\tilde{\mathbf{h}}]$. We then feed $\mathbf{h}$ into the classification layer to get the label estimation $l_i$ for each token in original input and train the model to minimize the cross entropy loss between $l_i$ and $y_i$. 
Note that we use identical demonstrations during training and testing, which is crucial for demonstration-based learning to work \citep{lee-etal-2022-good}.
\section{Pathological Demonstrations}
\label{sec:Pathological}

We refer to the demonstration constructed in Section~\ref{DemoNER} as  \textbf{ST}andard (\standard) demonstration. 
Suppose the model leverages the demonstrations in a human-analogous way and understands the meaning of them, there will be no more performance gains if we no longer provide correct example-label pairs or actual examples.
To this end, we design three pathological demonstrations by gradually removing such intuitively helpful information from \standard\  demonstrations: 
\begin{enumerate}
    \item \standardwrong\ (\textbf{S}tandard demonstration with \textbf{W}rong labels): Intuitively, the most helpful information in demonstrations is the correlation between provided examples and tag types, thus the first kind of pathological demonstrations provides wrong examples for each tag type on purpose.
    \item \standardnol\ (\textbf{S}tandard demonstration with \textbf{N}o label): 
    Furthermore, the existence of examples or tags in labeling part of the demonstration might give away hints, so we remove the labeling part to create the second pathological demonstration that consists of only contexts from the support set.
    \item \randomtotally\ (\textbf{T}otally \textbf{R}andom demonstration): Finally, we test a seemingly useless demonstration by using random token strings as demonstrations. Specifically, we replace every token in the demonstration \textbf{SN} with random tokens sampled from the vocabulary.
\end{enumerate}

We show templates and examples for these pathological demonstrations modified from standard demonstration for NER in Table~\ref{tab:templates}.

\begin{table}[t]
\resizebox{.48\textwidth}{!}{
\begin{tabular}{@{}lll@{}}
\toprule
Mode            & Template T                      & Example (for type PER)                                    \\ \midrule
\standard        & [SEP] $s\ e\ \text{is}\ t.$ & {[}SEP{]} Obama returns to White House. Obama is PER.  \\ \midrule
\standardwrong & [SEP] $s\ e\ \text{is}\ t'.$ & {[}SEP{]} Obama returns to White House. Obama is LOC.  \\ 
\standardnol & [SEP] $s$  & {[}SEP{]} Obama returns to White House.  \\ 
\randomtotally & [SEP] $s'$  & {[}SEP{]} similar Requiem tracking Michelle seeds 15th \\ \midrule
\randomsupport & [SEP] $s''$  & {[}SEP{]} war House to on Arafat returns               \\ \bottomrule
\end{tabular}}
\caption{\textbf{Templates and examples for different modes of demonstrations}. Here, $e$ denotes the entity example sampled for entity type $t$, and $s$ denotes the sentence from the support set that contains entity $e$ as type of $t$. $s', s''$ refer to the same sentence but with every token of it being replaced by random tokens sampled from whole vocabulary or \textit{relevant} vocabulary (Section~\ref{sec:SR}) respectively. All examples above are modified from the standard demonstration shown in Figure~\ref{fig:construction} and only part of them are displayed here. We show a list of real demonstrations we constructed and used in Appendix~\ref{app:example}.}
\label{tab:templates}
\end{table}

\section{Experiments}

\subsection{Few-Shot Datasets}
\paragraph{Datasets}
We conduct experiments on two sequence labeling tasks: named entity recognition and chunking. For NER task, we use dataset \textbf{CoNLL03}~\citep{tjong-kim-sang-de-meulder-2003-introduction}, and \textbf{OntoNotes 5.0}~\citep{weischedel2013ontonotes}. Since we primarily focus on named entities, we omit the 7 value types in OntoNotes following \citet{DBLP:journals/corr/abs-2109-13532}. In addition, we use \textbf{CoNLL00}~\citep{tjong-kim-sang-buchholz-2000-introduction} for the chunking task. Since the number of some phrase types is very limited, we only consider 6 most frequent types (which are \textit{NP, VP, PP, ADVP, SBAR} and \textit{ADJP}, accounting for 99\% of the labeled chunks).
\paragraph{Few-shot data sampling}
Different from sentence-level few-shot tasks, in sequence labeling, one sample for a class refers to a span in the sentence, and one sentence may contain multiple samples of different types. We follow the greedy sampling strategy proposed by \citet{yang-katiyar-2020-simple} to sample $K$ shots for each type in an increasing order with respect to their frequencies, the detailed algorithm can be found at Appendix~\ref{app:greedy_sampling}.
The detailed dataset statistics are shown in Table~\ref{tab:data}.

\begin{table}[t]
\begin{tabular}{@{}lrrrr@{}}
\toprule
Dataset         & $|Y|$ & L     & $|D_{support}|$     & $|D_{test}|$        \\ \midrule
CoNLL03         & 4     & 18    & 8.0\textsubscript{$\pm$1.1}    & 3453          \\
OntoNotes 5.0   & 11    & 21    & 26.6\textsubscript{$\pm$1.2}    & 12217         \\
CoNLL00         & 6     & 36    & 8.6\textsubscript{$\pm$0.8}    & 2012  \\\bottomrule
\end{tabular}
\caption{\textbf{Data Statistics}. $|Y|$: \# of entity types. L: average \# of tokens in input sentence. $|D_{support}|$: average \# of sentences in 5-shot support set over 5 different sub-samples. $|D_{test}|$: \# of sentences in test set.}
\label{tab:data}
\end{table}

\subsection{Implementation Details}
We use \texttt{bert-base-cased} model from HuggingFace \citep{wolf-etal-2020-transformers} as our backbone for all the experiments and set the batch size and learning rate to 4 and 2e-5, respectively, following \citet{lee-etal-2022-good}. We use NVIDIA GeForce RTX 3080 Ti to conduct all experiments.
For each variant, we run 50 epochs over 5 different sub-samples and 3 random seeds with early-stopping 20 and report its micro-F1 scores along with its recall and precision.

\begin{table*}[t]
\centering
\scalebox{0.78}{
	\begin{tabular}{lccccccccc}
        \toprule
        \multirow{3}{*}{\textbf{Mode}} &
        \multicolumn{6}{c}{\textbf{NER}} & \multicolumn{3}{c}{\textbf{Chunking}} \\
        \cmidrule(lr){2-7} \cmidrule(lr){8-10}
         & \multicolumn{3}{c}{\textbf{CoNLL03}} & \multicolumn{3}{c}{\textbf{OntoNotes 5.0}} & \multicolumn{3}{c}{\textbf{CoNLL00}}\\
        \cmidrule(lr){2-4} \cmidrule(lr){5-7} \cmidrule(lr){8-10} & 
        \textbf{F1} & \textbf{Precision} & \textbf{Recall} & 
        \textbf{F1} & \textbf{Precision} & \textbf{Recall} & 
        \textbf{F1} & \textbf{Precision} & \textbf{Recall} \\
        \midrule
        \textbf{NO} & 28.71\textsubscript{$\pm$10.31} & 39.96\textsubscript{$\pm$11.25} & 22.68\textsubscript{$\pm$9.09} & 37.37\textsubscript{$\pm$7.58} & 33.80\textsubscript{$\pm$6.79} & 41.92\textsubscript{$\pm$8.85} & 63.17\textsubscript{$\pm$4.22} & 59.28\textsubscript{$\pm$5.05} & 67.72\textsubscript{$\pm$3.51} \\
        \midrule
        \standard & 46.25\textsubscript{$\pm$5.41} & 47.92\textsubscript{$\pm$5.91} & 45.02\textsubscript{$\pm$6.06} & 40.21\textsubscript{$\pm$7.65} & 32.51\textsubscript{$\pm$6.87} & 52.82\textsubscript{$\pm$8.28} & 70.55\textsubscript{$\pm$3.08} & 66.53\textsubscript{$\pm$4.40} & 75.21\textsubscript{$\pm$2.11}\\
        \midrule
        \standardwrong & 46.23\textsubscript{$\pm$5.63} & 47.91\textsubscript{$\pm$6.04} & 45.01\textsubscript{$\pm$6.29} & 39.94\textsubscript{$\pm$7.38} & 32.27\textsubscript{$\pm$6.59} & 52.50\textsubscript{$\pm$8.11} & 70.75\textsubscript{$\pm$3.05} & 66.80\textsubscript{$\pm$4.39} & 75.33\textsubscript{$\pm$2.14}\\
        \standardnol & 45.74\textsubscript{$\pm$6.52} & 47.86\textsubscript{$\pm$6.23} & 44.31\textsubscript{$\pm$7.79} & 40.29\textsubscript{$\pm$6.76} & 32.46\textsubscript{$\pm$5.81} & 53.18\textsubscript{$\pm$8.10} & 69.94\textsubscript{$\pm$3.16} & 65.86\textsubscript{$\pm$4.48} & 74.70\textsubscript{$\pm$2.12}\\
        \randomtotally & 41.33\textsubscript{$\pm$7.36} & 45.41\textsubscript{$\pm$7.37} & 38.22\textsubscript{$\pm$7.65} & 39.71\textsubscript{$\pm$7.56} & 32.28\textsubscript{$\pm$6.56} & 51.63\textsubscript{$\pm$8.75} & 69.28\textsubscript{$\pm$2.78} & 64.75\textsubscript{$\pm$3.85} & 74.57\textsubscript{$\pm$1.66}\\
        \midrule
        \randomsupport & 45.99\textsubscript{$\pm$7.90} & 47.20\textsubscript{$\pm$7.84} & 45.09\textsubscript{$\pm$8.34} & 41.60\textsubscript{$\pm$7.05} & 33.96\textsubscript{$\pm$6.29} & 53.75\textsubscript{$\pm$7.80} & 70.63\textsubscript{$\pm$3.01} & 66.24\textsubscript{$\pm$4.29} & 75.75\textsubscript{$\pm$1.70}\\
        \bottomrule
    \end{tabular}
    }
    \caption{Main results for traditional token classification method (\textbf{NO}) and demonstration-based learning with different modes of demonstrations 
    under 5-shot scenario. 
    We report mean and standard deviation of F1-score, precision, and recall over 15 runs (5 different sub-samples and 3 random seeds).}
    \label{tab:results}
\end{table*}

\subsection{Results}

We show the detailed results for demonstration-based learning with standard demonstrations and pathological demonstrations in Section~\ref{sec:Pathological}, as well as traditional token classification methods with no demonstration in Table~\ref{tab:results}.

\paragraph{Demonstration is effective!}
Comparing results between no demonstration method (\textbf{NO}) and demonstration-based learning (\standard), we found that demonstrations improve the few-shot performance significantly (e.g., from 28.71 to 46.25 on CoNLL03, and from 63.17 to 70.55 on CoNLL00).
A closer look at reveals that the performance gains are mainly from a much higher recall for NER task, indicating demonstrations are mainly helping recognize more entities.

\paragraph{No need for labels?}
As shown in Table~\ref{tab:results}, there is no significant difference between the F1 scores of standard demonstration (\standard) and its pathological variation \standardwrong.
This suggests that a \emph{working} demonstration even does not need to have the correct labels. Moreover, even if we remove the entire labeling part (often perceived as the most important factor for demonstrations), the pathological demonstration \standardnol\ can still achieve as impressive results as \standard\ demonstration. Our results are consistent with a recent work \citep{min2022rethinking} that correct label mapping may not be needed for demonstrations to work, though we approach the robustness of demonstration based learning in a different classifier-based fine-tuning setting. 

\paragraph{Random demonstration also works.}
Surprisingly, there is significant performance gain when using demonstration \randomtotally\ over \textbf{NO} (e.g., from 28.71 to 41.33 on CoNLL03, and from 63.17 to 69.28 on CoNLL00), though the gap between \randomtotally\ and \standard\ demonstration still exists. Note that 
\randomtotally\ demonstration is no longer a real sentence and may not provide any meaningful or useful information. 

\subsection{Analysis}
\label{sec:SR}
This section provides a deeper understanding of why and how demonstration-based learning works given the counter-intuitive results in Section 5.3. 

\paragraph{Relevance of the tokens counts!}
Looking at the performance difference between demonstration \standardnol\ and \randomtotally, we hypothesize that a key factor might be the random tokens' relevance to the input sentence.
To test this, we construct a demonstration \randomsupport\ (\textbf{S}upport set sampled \textbf{R}andom demonstration) as shown in Table~\ref{tab:templates}, by first creating a  \textit{relevant vocabulary} consisting of only tokens appear in the support set \supportset, and then sampling tokens from this relevant vocabulary to replace tokens in demonstration. The result is shown in the last row of Table~\ref{tab:results}. We found that the performance of \randomsupport\ is superior to \randomtotally\ (45.99 v.s. 41.33 on CoNLL03) and comparable to \standard\ demonstration (45.99 v.s. 46.25 on CoNLL03). This implies that the relevance of tokens of the demonstration is very essential to demonstration-based learning.

\begin{figure}[t]
    \centerline{\includegraphics[width=0.46\textwidth]{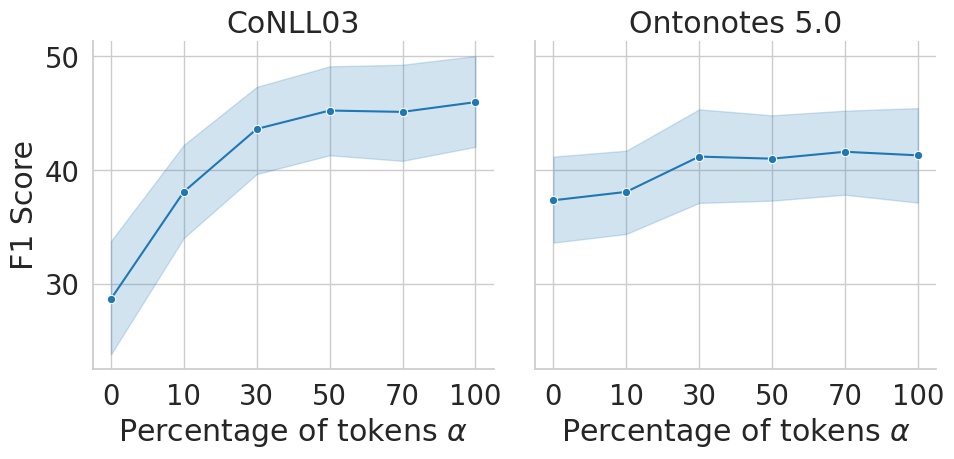}}
    \caption{\textbf{Impact of demonstrations length.} Results for traditional token classification methods ($\alpha = 0$) and demonstration-based learning with pathological demonstration \randomsupport\ of different length, where $\alpha$ denotes the length is $\alpha\%$ of the original \randomsupport\ demonstration.}
    \label{fig:length}
\end{figure}

\begin{figure}[t]
    \centering
    \includegraphics[width=0.4\textwidth]{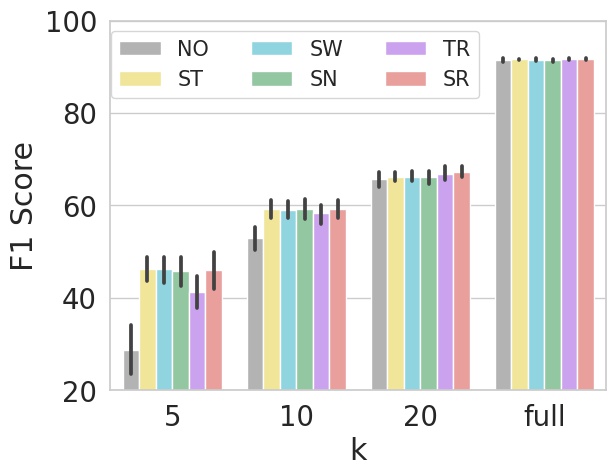}
    \caption{\textbf{Performance trends} under different level of data scarcity on CoNLL03 dataset.}
    \label{fig:trend}
\end{figure}

\paragraph{Length of demonstrations matters.}
Since there is not much semantic meaning included in the demonstration \randomsupport, another crucial difference between \randomsupport\ and \textbf{NO} is the length of random demonstrations.
Thus, we conducted ablation study by varying the length of demonstration \randomsupport. We evaluated the performance with demonstration consisting of $\alpha\%$ tokens of original \randomsupport\ demonstration.
As shown in Figure~\ref{fig:length}, the performance of demonstration improves from 28.71 to 45.99 on CoNLL03 and from 37.37 to 41.60 on OntoNotes 5.0 with the longer length of $\alpha$ from 0 to 100; it saturates (achieving 98\% of original \randomsupport\ demonstration's performance) at a relatively short length of $50\%$ of the original length. 
This suggests that a fair number of tokens is needed for demonstration \randomsupport\ to be working, and it seems  the length of demonstrations matters much more than their content.
Our finding here is consistent with the finding in \citet{xie2022an}.

\paragraph{The magic vanishes with more data.}
We further examine whether the performance gain of demonstration-based learning changes over different level of data scarcity, namely $K$-shots support set. We show results for the aforementioned (no) demonstrations under $K=5,10,20$ shots and full data in Figure~\ref{fig:trend}. The F1 score gain from demonstration is 17.54 (from 28.71 to 46.25) for 5-shot support set, 6.2 in the 10-shot setting, and negligible for the 20-shots support set and full data.
Consistent with \citet{lee-etal-2022-good, gao-etal-2021-making}, the performance gain (no matter standard or pathological) vanishes with more data. This indicates demonstrations have a strong boost on performance especially in extremely limited scenario, where there is no enough data for the model to fit well. 

\paragraph{Similar findings on Roberta.}
To see whether this counter-intuitive finding holds on other PLMs as well, we experimented on Roberta with \texttt{roberta-base} model from HuggingFace and show the results in Tabel~\ref{tab:roberta}. Similar to the results on BERT, standard demonstration improves the performance by 1.87 F1-score while pathological demonstrations with no intuitively meaningful information work as well. It implies that the cause behind this counter-intuitive finding is not only specific to BERT, but may aslo be prevalent with other PLMs.

\begin{table}[t]
\centering
\scalebox{0.9}{
	\begin{tabular}{lccc}
        \toprule
        \textbf{Mode} & \textbf{F1} & \textbf{Precision} & \textbf{Recall}\\
        \midrule
        \textbf{NO} & 52.08\textsubscript{$\pm$7.02} & 56.52\textsubscript{$\pm$6.46} & 48.42\textsubscript{$\pm$7.59} \\
        \midrule
        \textbf{\standard} & 53.95\textsubscript{$\pm$7.55} & 54.68\textsubscript{$\pm$7.80} & 53.36\textsubscript{$\pm$7.77} \\
        \textbf{\standardwrong} & 53.60\textsubscript{$\pm$7.41} & 54.33\textsubscript{$\pm$7.86} & 53.00\textsubscript{$\pm$7.36} \\
        \textbf{\standardnol} & 53.88\textsubscript{$\pm$7.21} & 54.80\textsubscript{$\pm$7.71} & 53.14\textsubscript{$\pm$7.36} \\
        \textbf{\randomtotally} & 53.57\textsubscript{$\pm$6.55} & 55.01\textsubscript{$\pm$6.92} & 52.25\textsubscript{$\pm$6.46} \\
        \midrule
        \textbf{\randomsupport} & 55.18\textsubscript{$\pm$6.23} & 55.77\textsubscript{$\pm$6.71} & 54.73\textsubscript{$\pm$6.43} \\
        \bottomrule
    \end{tabular}
    }
    \caption{Results for Roberta on CoNLL03 dataset under 5-shot scenario. 
    We report mean and standard deviation of F1-score, precision, and recall over 15 runs (5 different sub-samples and 3 random seeds).}
    \label{tab:roberta}
\end{table}
\section{Understanding the Demonstrations}
We take a closer look at the surprising performance of (pathological) demonstrations to examine whether such strong performance has any connections with spurious patterns or dataset bias, which the deep learning models are constantly being accused of leveraging \citep{wang2021measure}. As a case study, we 
use a carefully designed testbed NRB \citep{NRB} to diagnose Name Regularity Bias in the NER models learned with demonstrations (Section~\ref{sec:NRB}). 
We also conduct experiments on the more popular way of utilizing demonstrations with prompts  in Section~\ref{sec:LM-BFF} to show the effectiveness of pathological demonstrations. 

\subsection{Name Regularity Bias}
\label{sec:NRB}

Name Regularity Bias \citep{NRB, lin-etal-2020-rigorous} in NER occurs when a model relies on a signal from the entity name to make predictions and disregards evidence from the local context. For example, given the input sentence "\textit{Obama is located in far southwestern Fukui Prefecture.}", modern NER models may wrongly predict the entity \textit{Obama} as \textit{PER}, while the context clearly signals it is a \textit{LOC}. Therefore, \citet{NRB} carefully designed a testbed utilizing the Wikipedia disambiguation page to diagnose Name Regularity Bias of NER models. The NRB dataset contains 
examples whose labels can be easily inferred from the local context but hard to be tagged by a popular NER system. The WTS dataset is a domain control set that contains the  same query terms covered by NRB, but can be correctly labeled by both the popular NER tagger and local context-only tagger.

\subsubsection{Results}
\begin{figure}[t]
    \centering
    \includegraphics[width=0.46\textwidth]{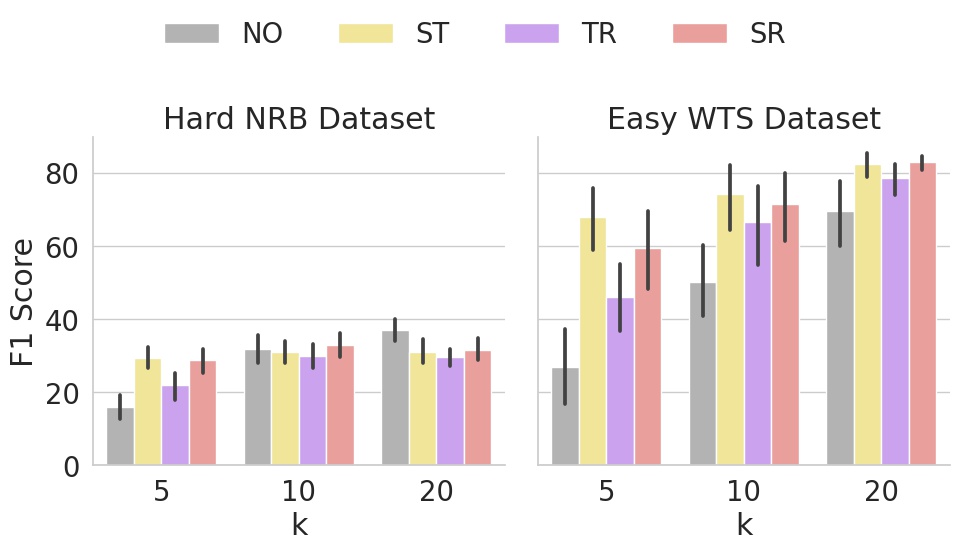}
    \caption{\textbf{Performance trend on NRB and WTS dataset}, with the model trained on the CoNLL03 dataset. 
    Though demonstration-based learning consistently brings performance gain on the WTS dataset (right), it helps less or hurt performance on the NRB dataset (left).}
    \label{fig:NRB}
\end{figure}

We use both the NRB and WTS datasets to evaluate the model trained with different modes of demonstrations on CoNLL03 under 5,10,20-shots support set and full data. The result is shown in Figure~\ref{fig:NRB}. As we can see, demonstration-based learning on the control set WTS consistently brings impressive performance gains under all low-resource settings (e.g. from 27.00 to 68.07 with 5-shots support set, and from 69.76 to 82.57 with 20-shots support set). On challenging dataset NRB, it only shows a performance gain from 15.82 to 29.44 with 5-shots support set, 
with no performance gain with 10-shots support set and even decreased F1 scores with the 20-shots support set. 
This suggests that demonstration-based learning leverages the name regularity bias to recognize entities rather than the context information. 

\subsubsection{Analysis}

\begin{table*}[h!]
\centering
\scalebox{0.78}{
	\begin{tabular}{llcccccccc}
        \toprule
        \multirow{2}{*}{\textbf{Dataset}} & \multirow{2}{*}{\textbf{Input}} &
        \multicolumn{4}{c}{\textbf{Confidence with mode \textbf{NO}}} &  \multicolumn{4}{c}{\textbf{Confidence with mode \standard}} \\
        \cmidrule(lr){3-6} \cmidrule(lr){7-10} & &
        \textbf{O} & \textbf{LOC} & \textbf{ORG} & \textbf{PER} &
        \textbf{O} & \textbf{LOC} & \textbf{ORG} & \textbf{PER} \\
        \midrule
        \multirow{2}{*}{\textbf{NRB}} & 
        A \colorbox{yellow}{post} office operated at Clinton from 1856 to 1859. &
        \textbf{\underline{0.88}} & 0.05 & 0.04 & 0.03 & \textbf{\underline{1.0}} & 0.0 & 0.0 & 0.0 \\
        & 
        A post office operated at \colorbox{yellow}{Clinton} from 1856 to 1859. &
        \textbf{0.37} & \underline{0.18} & 0.13 & 0.32 & 0.16 & \underline{0.33} & 0.13 & \textbf{0.38} \\
        \midrule
        \multirow{2}{*}{\textbf{WTS}} & 
        It was confirmed that \colorbox{yellow}{Clinton} had signed to 
        &
        0.32 & 0.10 & 0.15 & \textbf{\underline{0.43}} & 0.04 & 0.01 & 0.01 & \textbf{\underline{0.94}} \\
        & 
        \colorbox{yellow}{Clinton} has eight CDs and two DVDs available . &
        \textbf{0.67} & 0.05 & 0.09 & \underline{0.19} & 0.20 & 0.06 & 0.06 & \textbf{\underline{0.68}} \\
        \bottomrule
    \end{tabular}
    }
    \caption{\textbf{Examples from NRB and WTS dataset.} The token being predicted in the input is highlighted in \colorbox{yellow}{yellow}. The confidence score is \textbf{bold} for the final prediction (a.k.a the highest), and \underline{underlined} for the true label.}
    \label{tab:examples}
\end{table*}

\begin{figure}[ht]
    \centering
    \subfigure[NRB]{
        \includegraphics[width=0.45\textwidth]{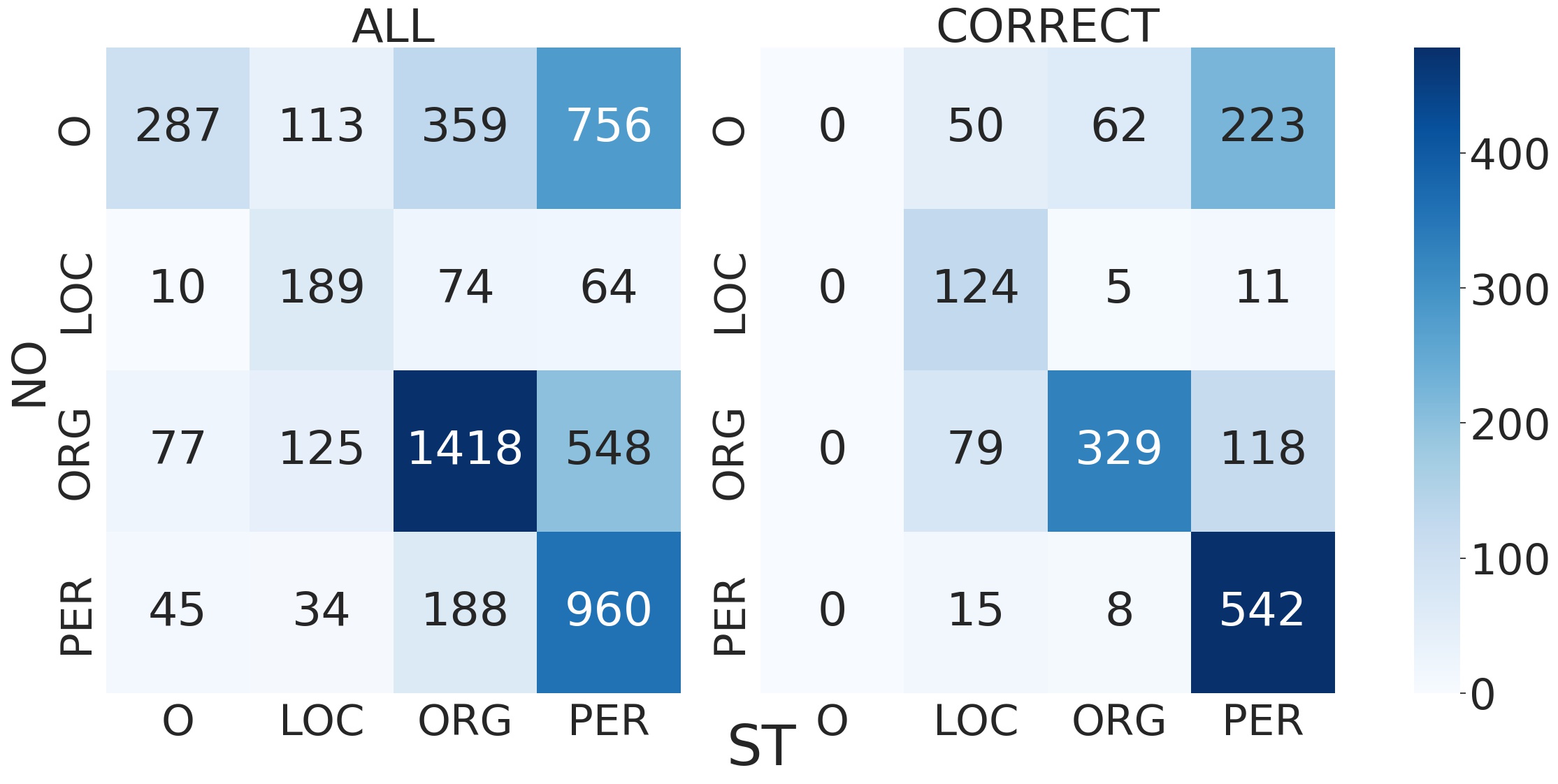}
        \label{fig:heat_NRB}
    }
    \subfigure[WTS]{
        \includegraphics[width=0.45\textwidth]{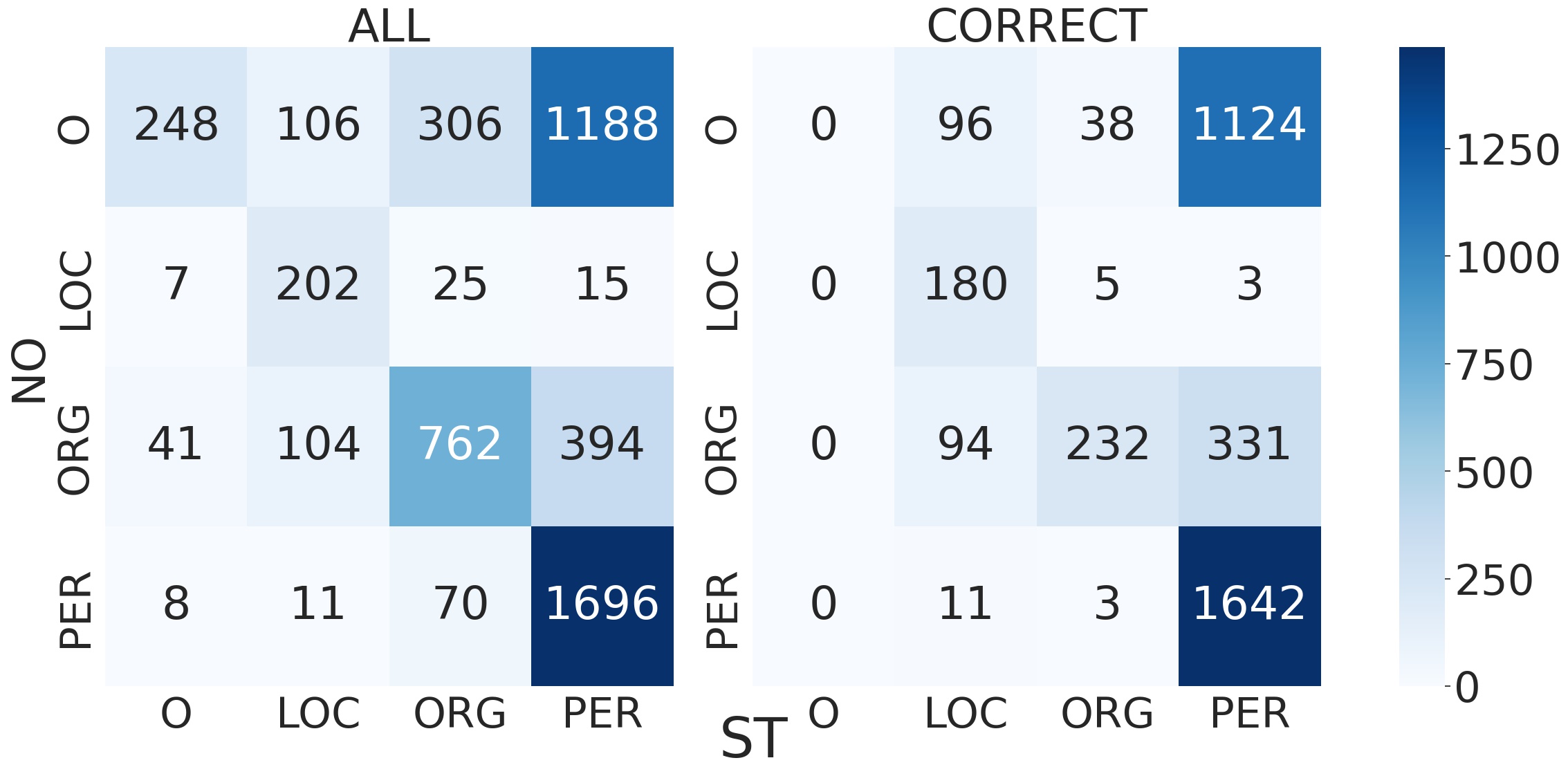}
        \label{fig:heat_WTS}
    }
    \caption{Number of prediction flips for entities from mode \textbf{NO} (row) to \standard\ (column). Left shows all prediction flips for entities while right shows the only ones that are correct with mode \standard. 
    }
    \label{fig:heat}
\end{figure}

\paragraph{Demonstrations bring no robust improvements}
To have a better understanding on how demonstrations affect the performance, we show the detailed prediction flips for all entities after adding \standard\ demonstrations to \textbf{NO}
in Figure~\ref{fig:heat}, where each cell shows the number of predictions that flip from the original prediction (row) to the new prediction with \standard\ demonstrations (column), while the diagonal represents the number of predictions that remain unchanged.
The left figure contains the overall number of such prediction flips and the right figure contains the number of such prediction flips that are correct.
As we can see, though there is a similar pattern of prediction flip on both NRB and WTS, the correctness of these prediction flips are different. For the challenging dataset NRB, the prediction flip with the highest number, namely O $\rightarrow$ PER and ORG $\rightarrow$ PER, have a relatively low correct ratio of only $29\%(223/756)$ and $22\%(118/548)$, compared with $95\%(1124/1188)$ and $84\%(331/394)$ respectively on WTS. The second row in Table~\ref{tab:examples} shows an example of the wrong prediction flip for token "\texttt{Clinton}" from \textit{O} to \textit{PER}, while the true label should be \textit{LOC}.
Demonstrations can better recognize entities, but can not distinguish well among the entity types and misguide the model to make more false positive predictions on NRB.
This further supports that 
demonstrations are not making robust improvements and simply leverage the name regularity bias.

\begin{figure}[ht]
    \centerline{\includegraphics[width=0.5\textwidth]{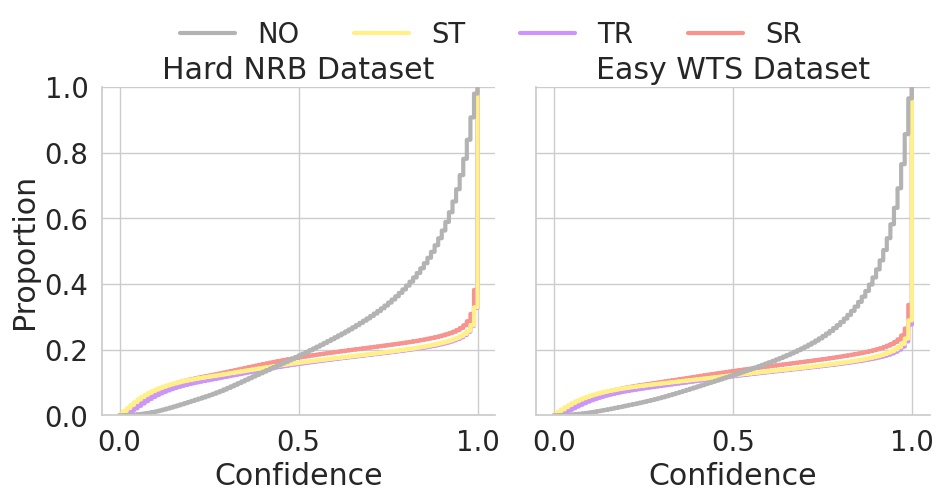}}
    \caption{\textbf{Empirical cumulative distribution function of model's  confidence on its final prediction.} 
    }
    \label{fig:confident}
\end{figure}

\paragraph{Demonstrations increase the confidence of model predictions}
To take a closer look at how these prediction flips work, we show the detailed confidence score (\textit{a.k.a} the probability) for models' predictions with some illustrative examples in Table~\ref{tab:examples}. Notably, in the first and third row, the confidence for token "\texttt{post}" to be \textit{O} and "\texttt{Clinton}" to be \textit{PER} increase from 0.88 to 1.0 and 0.43 to 0.94 respectively, therefore we hypothesize demonstrations increase the confidence of model's final prediction. We show empirical cumulative distribution function (ecdf) of the model's confidence for the final prediction with different modes of demonstrations on both NRB and WTS benchmarks in Figure~\ref{fig:confident}. We found that, with demonstrations (either standard or random), the model tends to be more confident.

\subsection{Demonstrations with prompt-tuning}
\label{sec:LM-BFF}
\begin{figure}[t]
    \centerline{\includegraphics[width=0.5\textwidth]{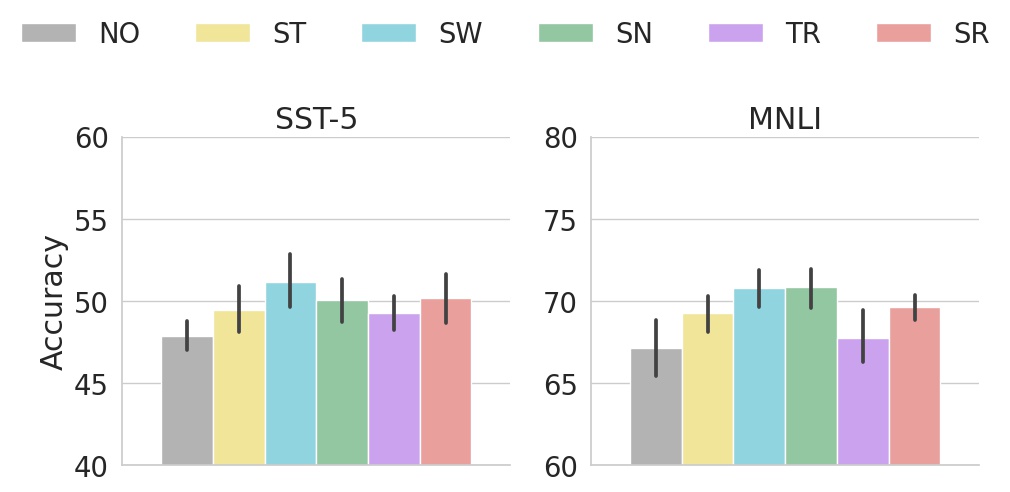}}
    \caption{\textbf{Results of using pathological demonstrations with LM-BFF on SST-5 and MNLI.} Here we report averaged accuracy over 5 random seeds.}
    \label{fig:LM-BFF}
\end{figure}

Our construction of pathological demonstrations can be easily generalized to other types of demonstration-based learning, such as the prompt-based fine-tuning used in LM-BFF \citep{gao-etal-2021-making}. Following their settings, we conduct experiments with \texttt{roberta-large} model on single-sentence task SST-5 \citep{socher-etal-2013-recursive} and sentence-pair task MNLI \citep{williams-etal-2018-broad} with 16-shots support set, as shown in Figure~\ref{fig:LM-BFF}. We found that the performance of pathological demonstrations is competitive with standard demonstrations,  consistent with our findings on sequence labeling tasks with classifier-based fine-tuning.

\section{Discussion and Conclusion}
In this paper, we study the robustness of demonstration-based learning by designing pathological demonstrations.
We found that, replacing demonstrations with random tokens still makes the model a better few-shot learner;  
the length of random token strings and the sampling space for random tokens are the main factors affecting the performance;
and demonstrations increase the confidence of model predictions on captured superficial patterns such as the name regularity bias.
Below we discuss the broader impacts of our findings.

Our findings imply that natural language is not sufficiently understood by the PLMs when being utilized together with demonstrations, since random token strings also lead to similar strong performances.
Similar to our work, recent studies \citep{ri-tsuruoka-2022-pretraining, chiang2022transferability} also found models pretrained on random token strings with a nesting dependency structure still provide transferable knowledge to downstream fine-tuning, suggesting the insufficient utilization of natural language during pretraining. 
We urge future work on demonstration based learning to think twice about their model performance gains by designing more robustness tests and ablation studies. 

Our work also calls for a better design of  demonstrations that are free of spurious patterns,  
as existing demonstrations are less effective while there is no spurious patterns to leverage (see Section~\ref{sec:NRB}).
Future work should not only optimize the performances of demonstration based methods on the validation set, but also pay attention to whether these models suffer from spurious patterns and how to increase their generalization abilities.  

\section*{Limitations}

This work is subject to several limitations.
First, we primarily look at sequence labeling tasks in this work, and have not applied similar techniques for other tasks such as text classification.
Second, we followed \citet{lee-etal-2022-good} to use the widely used  \texttt{bert-base-cased} as our backbone for most of our experiments.  
A thorough examination of other language models such as T5 is needed, which we leave as future work. 
Finally, the present work focuses on understanding the robustness of demonstration-based learning, and we have not developed any practical guidelines on how to use our findings to design more effective demonstrations.

\section*{Acknowledgements} The authors would like to thank reviewers for their helpful insights and feedback. This work is funded in part by grants from Adobe and Amazon. 

\bibliography{anthology,custom}

\begin{thebibliography}{40}
\expandafter\ifx\csname natexlab\endcsname\relax\def\natexlab#1{#1}\fi

\bibitem[{Brown et~al.(2020)Brown, Mann, Ryder, Subbiah, Kaplan, Dhariwal,
  Neelakantan, Shyam, Sastry, Askell, Agarwal, Herbert{-}Voss, Krueger,
  Henighan, Child, Ramesh, Ziegler, Wu, Winter, Hesse, Chen, Sigler, Litwin,
  Gray, Chess, Clark, Berner, McCandlish, Radford, Sutskever, and
  Amodei}]{NEURIPS2020_1457c0d6}
Tom~B. Brown, Benjamin Mann, Nick Ryder, Melanie Subbiah, Jared Kaplan,
  Prafulla Dhariwal, Arvind Neelakantan, Pranav Shyam, Girish Sastry, Amanda
  Askell, Sandhini Agarwal, Ariel Herbert{-}Voss, Gretchen Krueger, Tom
  Henighan, Rewon Child, Aditya Ramesh, Daniel~M. Ziegler, Jeffrey Wu, Clemens
  Winter, Christopher Hesse, Mark Chen, Eric Sigler, Mateusz Litwin, Scott
  Gray, Benjamin Chess, Jack Clark, Christopher Berner, Sam McCandlish, Alec
  Radford, Ilya Sutskever, and Dario Amodei. 2020.
\newblock \href
  {https://proceedings.neurips.cc/paper/2020/hash/1457c0d6bfcb4967418bfb8ac142f64a-Abstract.html}
  {Language models are few-shot learners}.
\newblock In \emph{Advances in Neural Information Processing Systems 33: Annual
  Conference on Neural Information Processing Systems 2020, NeurIPS 2020,
  December 6-12, 2020, virtual}.

\bibitem[{Chiang and Lee(2022)}]{chiang2022transferability}
Cheng-Han Chiang and Hung-yi Lee. 2022.
\newblock On the transferability of pre-trained language models: A study from
  artificial datasets.
\newblock In \emph{Proceedings of the AAAI Conference on Artificial
  Intelligence}, volume~36, pages 10518--10525.

\bibitem[{Devlin et~al.(2019)Devlin, Chang, Lee, and
  Toutanova}]{devlin-etal-2019-bert}
Jacob Devlin, Ming-Wei Chang, Kenton Lee, and Kristina Toutanova. 2019.
\newblock \href {https://doi.org/10.18653/v1/N19-1423} {{BERT}: Pre-training of
  deep bidirectional transformers for language understanding}.
\newblock In \emph{Proceedings of the 2019 Conference of the North {A}merican
  Chapter of the Association for Computational Linguistics: Human Language
  Technologies, Volume 1 (Long and Short Papers)}, pages 4171--4186,
  Minneapolis, Minnesota. Association for Computational Linguistics.

\bibitem[{Gao et~al.(2021)Gao, Fisch, and Chen}]{gao-etal-2021-making}
Tianyu Gao, Adam Fisch, and Danqi Chen. 2021.
\newblock \href {https://doi.org/10.18653/v1/2021.acl-long.295} {Making
  pre-trained language models better few-shot learners}.
\newblock In \emph{Proceedings of the 59th Annual Meeting of the Association
  for Computational Linguistics and the 11th International Joint Conference on
  Natural Language Processing (Volume 1: Long Papers)}, pages 3816--3830,
  Online. Association for Computational Linguistics.

\bibitem[{Garg et~al.(2022)Garg, Tsipras, Liang, and
  Valiant}]{garg2022transformers}
Shivam Garg, Dimitris Tsipras, Percy Liang, and Gregory Valiant. 2022.
\newblock \href {http://arxiv.org/abs/2208.01066} {What can transformers learn
  in-context? a case study of simple function classes}.

\bibitem[{Ghaddar et~al.(2021)Ghaddar, Langlais, Rashid, and
  Rezagholizadeh}]{NRB}
Abbas Ghaddar, Philippe Langlais, Ahmad Rashid, and Mehdi Rezagholizadeh. 2021.
\newblock \href {https://doi.org/10.1162/tacl_a_00386} {{Context-aware
  Adversarial Training for Name Regularity Bias in Named Entity Recognition}}.
\newblock \emph{Transactions of the Association for Computational Linguistics},
  9:586--604.

\bibitem[{Huang et~al.(2021)Huang, Li, Subudhi, Jose, Balakrishnan, Chen, Peng,
  Gao, and Han}]{huang-etal-2021-shot}
Jiaxin Huang, Chunyuan Li, Krishan Subudhi, Damien Jose, Shobana Balakrishnan,
  Weizhu Chen, Baolin Peng, Jianfeng Gao, and Jiawei Han. 2021.
\newblock \href {https://doi.org/10.18653/v1/2021.emnlp-main.813} {Few-shot
  named entity recognition: An empirical baseline study}.
\newblock In \emph{Proceedings of the 2021 Conference on Empirical Methods in
  Natural Language Processing}, pages 10408--10423, Online and Punta Cana,
  Dominican Republic. Association for Computational Linguistics.

\bibitem[{Kim et~al.(2022)Kim, Kim, Cho, Jo, Lee, goo Lee, Yoo, and
  Kim}]{kim2022groundtruth}
Junyeob Kim, Hyuhng~Joon Kim, Hyunsoo Cho, Hwiyeol Jo, Sang-Woo Lee, Sang goo
  Lee, Kang~Min Yoo, and Taeuk Kim. 2022.
\newblock \href {http://arxiv.org/abs/2205.12685} {Ground-truth labels matter:
  A deeper look into input-label demonstrations}.

\bibitem[{Lake et~al.(2015)Lake, Salakhutdinov, and Tenenbaum}]{lake2015human}
Brenden Lake, Ruslan Salakhutdinov, and Joshua Tenenbaum. 2015.
\newblock \href {https://doi.org/10.1126/science.aab3050} {Human-level concept
  learning through probabilistic program induction}.
\newblock \emph{Science}, 350:1332--1338.

\bibitem[{Lee et~al.(2022)Lee, Kadakia, Tan, Agarwal, Feng, Shibuya, Mitani,
  Sekiya, Pujara, and Ren}]{lee-etal-2022-good}
Dong-Ho Lee, Akshen Kadakia, Kangmin Tan, Mahak Agarwal, Xinyu Feng, Takashi
  Shibuya, Ryosuke Mitani, Toshiyuki Sekiya, Jay Pujara, and Xiang Ren. 2022.
\newblock \href {https://doi.org/10.18653/v1/2022.acl-long.192} {Good examples
  make a faster learner: Simple demonstration-based learning for low-resource
  {NER}}.
\newblock In \emph{Proceedings of the 60th Annual Meeting of the Association
  for Computational Linguistics (Volume 1: Long Papers)}, pages 2687--2700,
  Dublin, Ireland. Association for Computational Linguistics.

\bibitem[{Lewis et~al.(2020)Lewis, Liu, Goyal, Ghazvininejad, Mohamed, Levy,
  Stoyanov, and Zettlemoyer}]{lewis-etal-2020-bart}
Mike Lewis, Yinhan Liu, Naman Goyal, Marjan Ghazvininejad, Abdelrahman Mohamed,
  Omer Levy, Veselin Stoyanov, and Luke Zettlemoyer. 2020.
\newblock \href {https://doi.org/10.18653/v1/2020.acl-main.703} {{BART}:
  Denoising sequence-to-sequence pre-training for natural language generation,
  translation, and comprehension}.
\newblock In \emph{Proceedings of the 58th Annual Meeting of the Association
  for Computational Linguistics}, pages 7871--7880, Online. Association for
  Computational Linguistics.

\bibitem[{Liang et~al.(2022)Liang, Zhang, Cheng, Bi, Zhang, Tan, Huang, Huang,
  and Chen}]{liang2022contrastive}
Xiaozhuan Liang, Ningyu Zhang, Siyuan Cheng, Zhen Bi, Zhenru Zhang, Chuanqi
  Tan, Songfang Huang, Fei Huang, and Huajun Chen. 2022.
\newblock Contrastive demonstration tuning for pre-trained language models.
\newblock \emph{arXiv preprint arXiv:2204.04392}.

\bibitem[{Lin et~al.(2020)Lin, Lu, Tang, Han, Sun, Wei, and
  Yuan}]{lin-etal-2020-rigorous}
Hongyu Lin, Yaojie Lu, Jialong Tang, Xianpei Han, Le~Sun, Zhicheng Wei, and
  Nicholas~Jing Yuan. 2020.
\newblock \href {https://doi.org/10.18653/v1/2020.emnlp-main.592} {A rigorous
  study on named entity recognition: Can fine-tuning pretrained model lead to
  the promised land?}
\newblock In \emph{Proceedings of the 2020 Conference on Empirical Methods in
  Natural Language Processing (EMNLP)}, pages 7291--7300, Online. Association
  for Computational Linguistics.

\bibitem[{Liu et~al.(2021{\natexlab{a}})Liu, Shen, Zhang, Dolan, Carin, and
  Chen}]{liu2021makes}
Jiachang Liu, Dinghan Shen, Yizhe Zhang, Bill Dolan, Lawrence Carin, and Weizhu
  Chen. 2021{\natexlab{a}}.
\newblock What makes good in-context examples for gpt-$3 $?
\newblock \emph{arXiv preprint arXiv:2101.06804}.

\bibitem[{Liu et~al.(2021{\natexlab{b}})Liu, Yuan, Fu, Jiang, Hayashi, and
  Neubig}]{liu2021pre}
Pengfei Liu, Weizhe Yuan, Jinlan Fu, Zhengbao Jiang, Hiroaki Hayashi, and
  Graham Neubig. 2021{\natexlab{b}}.
\newblock \href {https://arxiv.org/abs/2107.13586} {Pre-train, prompt, and
  predict: A systematic survey of prompting methods in natural language
  processing}.
\newblock \emph{ArXiv preprint}, abs/2107.13586.

\bibitem[{Liu et~al.(2019)Liu, Ott, Goyal, Du, Joshi, Chen, Levy, Lewis,
  Zettlemoyer, and Stoyanov}]{DBLP:journals/corr/abs-1907-11692}
Yinhan Liu, Myle Ott, Naman Goyal, Jingfei Du, Mandar Joshi, Danqi Chen, Omer
  Levy, Mike Lewis, Luke Zettlemoyer, and Veselin Stoyanov. 2019.
\newblock \href {http://arxiv.org/abs/1907.11692} {Roberta: {A} robustly
  optimized {BERT} pretraining approach}.
\newblock \emph{CoRR}, abs/1907.11692.

\bibitem[{Logan~IV et~al.(2021)Logan~IV, Bala{\v{z}}evi{\'c}, Wallace, Petroni,
  Singh, and Riedel}]{logan2021cutting}
Robert~L Logan~IV, Ivana Bala{\v{z}}evi{\'c}, Eric Wallace, Fabio Petroni,
  Sameer Singh, and Sebastian Riedel. 2021.
\newblock Cutting down on prompts and parameters: Simple few-shot learning with
  language models.
\newblock \emph{arXiv preprint arXiv:2106.13353}.

\bibitem[{Lu et~al.(2021)Lu, Bartolo, Moore, Riedel, and
  Stenetorp}]{lu2021fantastically}
Yao Lu, Max Bartolo, Alastair Moore, Sebastian Riedel, and Pontus Stenetorp.
  2021.
\newblock Fantastically ordered prompts and where to find them: Overcoming
  few-shot prompt order sensitivity.
\newblock \emph{arXiv preprint arXiv:2104.08786}.

\bibitem[{Ma et~al.(2021)Ma, Zhou, Gui, Tan, Zhang, and
  Huang}]{DBLP:journals/corr/abs-2109-13532}
Ruotian Ma, Xin Zhou, Tao Gui, Yiding Tan, Qi~Zhang, and Xuanjing Huang. 2021.
\newblock \href {http://arxiv.org/abs/2109.13532} {Template-free prompt tuning
  for few-shot {NER}}.
\newblock \emph{CoRR}, abs/2109.13532.

\bibitem[{Min et~al.(2022{\natexlab{a}})Min, Lewis, Hajishirzi, and
  Zettlemoyer}]{min-etal-2022-noisy}
Sewon Min, Mike Lewis, Hannaneh Hajishirzi, and Luke Zettlemoyer.
  2022{\natexlab{a}}.
\newblock \href {https://doi.org/10.18653/v1/2022.acl-long.365} {Noisy channel
  language model prompting for few-shot text classification}.
\newblock In \emph{Proceedings of the 60th Annual Meeting of the Association
  for Computational Linguistics (Volume 1: Long Papers)}, pages 5316--5330,
  Dublin, Ireland. Association for Computational Linguistics.

\bibitem[{Min et~al.(2022{\natexlab{b}})Min, Lyu, Holtzman, Artetxe, Lewis,
  Hajishirzi, and Zettlemoyer}]{min2022rethinking}
Sewon Min, Xinxi Lyu, Ari Holtzman, Mikel Artetxe, Mike Lewis, Hannaneh
  Hajishirzi, and Luke Zettlemoyer. 2022{\natexlab{b}}.
\newblock Rethinking the role of demonstrations: What makes in-context learning
  work?
\newblock \emph{arXiv preprint arXiv:2202.12837}.

\bibitem[{Mishra et~al.(2021)Mishra, Khashabi, Baral, and
  Hajishirzi}]{mishra2021cross}
Swaroop Mishra, Daniel Khashabi, Chitta Baral, and Hannaneh Hajishirzi. 2021.
\newblock Cross-task generalization via natural language crowdsourcing
  instructions.
\newblock \emph{arXiv preprint arXiv:2104.08773}.

\bibitem[{Radford et~al.(2019)Radford, Wu, Child, Luan, Amodei, Sutskever
  et~al.}]{radford2019language}
Alec Radford, Jeffrey Wu, Rewon Child, David Luan, Dario Amodei, Ilya
  Sutskever, et~al. 2019.
\newblock Language models are unsupervised multitask learners.
\newblock \emph{OpenAI blog}, 1(8):9.

\bibitem[{Ri and Tsuruoka(2022)}]{ri-tsuruoka-2022-pretraining}
Ryokan Ri and Yoshimasa Tsuruoka. 2022.
\newblock \href {https://doi.org/10.18653/v1/2022.acl-long.504} {Pretraining
  with artificial language: Studying transferable knowledge in language
  models}.
\newblock In \emph{Proceedings of the 60th Annual Meeting of the Association
  for Computational Linguistics (Volume 1: Long Papers)}, pages 7302--7315,
  Dublin, Ireland. Association for Computational Linguistics.

\bibitem[{Schick and
  Sch{\"u}tze(2021{\natexlab{a}})}]{schick-schutze-2021-exploiting}
Timo Schick and Hinrich Sch{\"u}tze. 2021{\natexlab{a}}.
\newblock \href {https://doi.org/10.18653/v1/2021.eacl-main.20} {Exploiting
  cloze-questions for few-shot text classification and natural language
  inference}.
\newblock In \emph{Proceedings of the 16th Conference of the European Chapter
  of the Association for Computational Linguistics: Main Volume}, pages
  255--269, Online. Association for Computational Linguistics.

\bibitem[{Schick and
  Sch{\"u}tze(2021{\natexlab{b}})}]{schick-schutze-2021-just}
Timo Schick and Hinrich Sch{\"u}tze. 2021{\natexlab{b}}.
\newblock \href {https://doi.org/10.18653/v1/2021.naacl-main.185} {It{'}s not
  just size that matters: Small language models are also few-shot learners}.
\newblock In \emph{Proceedings of the 2021 Conference of the North American
  Chapter of the Association for Computational Linguistics: Human Language
  Technologies}, pages 2339--2352, Online. Association for Computational
  Linguistics.

\bibitem[{Socher et~al.(2013)Socher, Perelygin, Wu, Chuang, Manning, Ng, and
  Potts}]{socher-etal-2013-recursive}
Richard Socher, Alex Perelygin, Jean Wu, Jason Chuang, Christopher~D. Manning,
  Andrew Ng, and Christopher Potts. 2013.
\newblock \href {https://aclanthology.org/D13-1170} {Recursive deep models for
  semantic compositionality over a sentiment treebank}.
\newblock In \emph{Proceedings of the 2013 Conference on Empirical Methods in
  Natural Language Processing}, pages 1631--1642, Seattle, Washington, USA.
  Association for Computational Linguistics.

\bibitem[{Tjong Kim~Sang and
  Buchholz(2000)}]{tjong-kim-sang-buchholz-2000-introduction}
Erik~F. Tjong Kim~Sang and Sabine Buchholz. 2000.
\newblock \href {https://aclanthology.org/W00-0726} {Introduction to the
  {C}o{NLL}-2000 shared task chunking}.
\newblock In \emph{Fourth Conference on Computational Natural Language Learning
  and the Second Learning Language in Logic Workshop}.

\bibitem[{Tjong Kim~Sang and
  De~Meulder(2003)}]{tjong-kim-sang-de-meulder-2003-introduction}
Erik~F. Tjong Kim~Sang and Fien De~Meulder. 2003.
\newblock \href {https://aclanthology.org/W03-0419} {Introduction to the
  {C}o{NLL}-2003 shared task: Language-independent named entity recognition}.
\newblock In \emph{Proceedings of the Seventh Conference on Natural Language
  Learning at {HLT}-{NAACL} 2003}, pages 142--147.

\bibitem[{Utama et~al.(2021)Utama, Moosavi, Sanh, and
  Gurevych}]{utama-etal-2021-avoiding}
Prasetya Utama, Nafise~Sadat Moosavi, Victor Sanh, and Iryna Gurevych. 2021.
\newblock \href {https://doi.org/10.18653/v1/2021.emnlp-main.713} {Avoiding
  inference heuristics in few-shot prompt-based finetuning}.
\newblock In \emph{Proceedings of the 2021 Conference on Empirical Methods in
  Natural Language Processing}, pages 9063--9074, Online and Punta Cana,
  Dominican Republic. Association for Computational Linguistics.

\bibitem[{Wang et~al.(2021)Wang, Wang, and Yang}]{wang2021measure}
Xuezhi Wang, Haohan Wang, and Diyi Yang. 2021.
\newblock Measure and improve robustness in nlp models: A survey.
\newblock \emph{arXiv preprint arXiv:2112.08313}.

\bibitem[{Webson and Pavlick(2021)}]{webson2021prompt}
Albert Webson and Ellie Pavlick. 2021.
\newblock Do prompt-based models really understand the meaning of their
  prompts?
\newblock \emph{arXiv preprint arXiv:2109.01247}.

\bibitem[{Wei et~al.(2022)Wei, Wang, Schuurmans, Bosma, Chi, Le, and
  Zhou}]{wei2022chain}
Jason Wei, Xuezhi Wang, Dale Schuurmans, Maarten Bosma, Ed~Chi, Quoc Le, and
  Denny Zhou. 2022.
\newblock Chain of thought prompting elicits reasoning in large language
  models.
\newblock \emph{arXiv preprint arXiv:2201.11903}.

\bibitem[{Weischedel et~al.(2013)Weischedel, Palmer, Marcus, Hovy, Pradhan,
  Ramshaw, Xue, Taylor, Kaufman, Franchini et~al.}]{weischedel2013ontonotes}
Ralph Weischedel, Martha Palmer, Mitchell Marcus, Eduard Hovy, Sameer Pradhan,
  Lance Ramshaw, Nianwen Xue, Ann Taylor, Jeff Kaufman, Michelle Franchini,
  et~al. 2013.
\newblock Ontonotes release 5.0 ldc2013t19.
\newblock \emph{Linguistic Data Consortium, Philadelphia, PA}, 23.

\bibitem[{Williams et~al.(2018)Williams, Nangia, and
  Bowman}]{williams-etal-2018-broad}
Adina Williams, Nikita Nangia, and Samuel Bowman. 2018.
\newblock \href {https://doi.org/10.18653/v1/N18-1101} {A broad-coverage
  challenge corpus for sentence understanding through inference}.
\newblock In \emph{Proceedings of the 2018 Conference of the North {A}merican
  Chapter of the Association for Computational Linguistics: Human Language
  Technologies, Volume 1 (Long Papers)}, pages 1112--1122, New Orleans,
  Louisiana. Association for Computational Linguistics.

\bibitem[{Wolf et~al.(2020)Wolf, Debut, Sanh, Chaumond, Delangue, Moi, Cistac,
  Rault, Louf, Funtowicz, Davison, Shleifer, von Platen, Ma, Jernite, Plu, Xu,
  Le~Scao, Gugger, Drame, Lhoest, and Rush}]{wolf-etal-2020-transformers}
Thomas Wolf, Lysandre Debut, Victor Sanh, Julien Chaumond, Clement Delangue,
  Anthony Moi, Pierric Cistac, Tim Rault, Remi Louf, Morgan Funtowicz, Joe
  Davison, Sam Shleifer, Patrick von Platen, Clara Ma, Yacine Jernite, Julien
  Plu, Canwen Xu, Teven Le~Scao, Sylvain Gugger, Mariama Drame, Quentin Lhoest,
  and Alexander Rush. 2020.
\newblock \href {https://doi.org/10.18653/v1/2020.emnlp-demos.6} {Transformers:
  State-of-the-art natural language processing}.
\newblock In \emph{Proceedings of the 2020 Conference on Empirical Methods in
  Natural Language Processing: System Demonstrations}, pages 38--45, Online.
  Association for Computational Linguistics.

\bibitem[{Xie et~al.(2020)Xie, Dai, Hovy, Luong, and Le}]{XieSemi}
Qizhe Xie, Zihang Dai, Eduard Hovy, Thang Luong, and Quoc Le. 2020.
\newblock \href
  {https://proceedings.neurips.cc/paper/2020/file/44feb0096faa8326192570788b38c1d1-Paper.pdf}
  {Unsupervised data augmentation for consistency training}.
\newblock In \emph{Advances in Neural Information Processing Systems},
  volume~33, pages 6256--6268. Curran Associates, Inc.

\bibitem[{Xie et~al.(2022)Xie, Raghunathan, Liang, and Ma}]{xie2022an}
Sang~Michael Xie, Aditi Raghunathan, Percy Liang, and Tengyu Ma. 2022.
\newblock \href {https://openreview.net/forum?id=RdJVFCHjUMI} {An explanation
  of in-context learning as implicit bayesian inference}.
\newblock In \emph{International Conference on Learning Representations}.

\bibitem[{Yang and Katiyar(2020)}]{yang-katiyar-2020-simple}
Yi~Yang and Arzoo Katiyar. 2020.
\newblock \href {https://doi.org/10.18653/v1/2020.emnlp-main.516} {Simple and
  effective few-shot named entity recognition with structured nearest neighbor
  learning}.
\newblock In \emph{Proceedings of the 2020 Conference on Empirical Methods in
  Natural Language Processing (EMNLP)}, pages 6365--6375, Online. Association
  for Computational Linguistics.

\bibitem[{Zhao et~al.(2021)Zhao, Wallace, Feng, Klein, and
  Singh}]{zhao2021calibrate}
Zihao Zhao, Eric Wallace, Shi Feng, Dan Klein, and Sameer Singh. 2021.
\newblock Calibrate before use: Improving few-shot performance of language
  models.
\newblock In \emph{International Conference on Machine Learning}, pages
  12697--12706. PMLR.

\end{thebibliography}
\bibliographystyle{acl_natbib}

\clearpage
\appendix

\begin{table*}[!ht]
\begin{tabular}{lp{0.9\textwidth}}
\toprule
\standard
& [SEP] 9/16 - Luo Yigang ( China ) beat Jason Wong ( Malaysia ) 15-5 15-6 China is LOC . [SEP] Fox said the British government wanted an end to the alleged harassment of its nationals at Dhaka airport by customs officials . British is MISC . [SEP] One dealer said positive stances from Merrill Lynch and SBC Warburg were the key factors behind the gains . Merrill Lynch is ORG . [SEP] +2 D.A. Weibring through 12 D.A. Weibring is PER . \\ \midrule
\standardwrong &  [SEP] 9/16 - Luo Yigang ( China ) beat Jason Wong ( Malaysia ) 15-5 15-6 China is MISC . [SEP] Fox said the British government wanted an end to the alleged harassment of its nationals at Dhaka airport by customs officials . British is ORG . [SEP] One dealer said positive stances from Merrill Lynch and SBC Warburg were the key factors behind the gains . Merrill Lynch is PER . [SEP] +2 D.A. Weibring through 12 D.A. Weibring is LOC .  \\ \midrule
\standardnol &  [SEP] 9/16 - Luo Yigang ( China ) beat Jason Wong ( Malaysia ) 15-5 15-6 [SEP] Fox said the British government wanted an end to the alleged harassment of its nationals at Dhaka airport by customs officials . [SEP] One dealer said positive stances from Merrill Lynch and SBC Warburg were the key factors behind the gains . [SEP] +2 D.A. Weibring through 12 \\ \midrule
\randomtotally &  [SEP] \#\#llan costs similar Requiem tracking Michelle seeds 15th HM influenced \#\#OH \#\#inia canyon visited USB punished \#\#ter hadn mom \#\#BA \#\#rrow \#\#hetto \#\#loss idea [SEP] carriages \#\#uk Mellon inconsistent archaeologists Server quartet Low Downs \#\#izations Bears \#\#titis again falsely sprawling Dennis hey plural exam goalkeeper kingdom Argentine [SEP] befriended \#\#ndi accept 1926 symbolic Colonel reviewer sketch rabbi Tampa \#\#orra tour Jul minorities \#\#iary closing Beta Sunday Jai counts quasi \#\#uminous [SEP] ambitious Funk Got \#\#orm types Another Elements growled \#\#aris evaluation resulted \\ \midrule
\randomsupport &  [SEP] Everton Merrill gains \#\#s One Moldova Ho beauty British qualifier S Lynch Dhaka through said 1995 Merrill \#\#ull beauty opening 12 working 9 . [SEP] Ta qualifier of 9 Russian through harassment Ho Dhaka up England airport its key \#\#burg republic \#\#man \#\#nch called Malaysia wounds ) [SEP] by \#\#ron China \#\#burg dealer ( Malaysia said Glenn up 9 in customs \#\#tars officials at + factors Jason Tale \#\#nife \#\#s [SEP] \#\#tars taken up behind husband 12 end Yi dealer S government \\ \bottomrule
\end{tabular}
\caption{\textbf{Example demonstrations for different modes constructed with method in Section~\ref{sec:Pathological}}. The dataset used here is CoNLL03 for NER task. Example demonstrations for \randomtotally\ and \randomsupport\ are shown as a string of tokens.}
\label{tab:demo_example_ner}
\end{table*}

\section{Additional Experimental Details}
We use \texttt{bert-base-cased} model from HuggingFace \citep{wolf-etal-2020-transformers} as our backbone and use NVIDIA GeForce RTX 3080 Ti to conduct all experiments. The model have roughly 110M parameters and takes 1 hour for each mode of demonstration on average to train under 5-shot scenario.

\section{Example Demonstrations}
\label{app:example}
We show a list of real demonstrations we constructed and used in the experiments for CoNLL03 and CoNLL00 in Table~\ref{tab:demo_example_ner} and Table~\ref{tab:demo_example_cnk}.

\section{Sampling Algorithm}
\label{app:greedy_sampling}

We follow the greedy sampling strategy proposed by \citet{yang-katiyar-2020-simple} to sample $K$ shots for each tag in an increasing order with respect to their frequencies, the detailed algorithm is shown in Algorithm~\ref{algo:greedy}.

\begin{algorithm}[h]
\caption{Greedy sampling}
\label{algo:greedy}
\begin{algorithmic}[1]
\REQUIRE \# of shot $K$, labeled set $\mathbf{X}$ with tag set $\mathcal{C}$
\STATE Sort classes in $\mathcal{C}$ based on their freq. in $\mathbf{X}$
\STATE $S \gets \phi $ //Initialize the support set
\STATE $\{ \text{Count}_i \gets 0 \}$ //Initialize counts of entity classes in $\mathcal{S}$
\WHILE {$i < |\mathcal{C}|$}
\WHILE{$\text{Count}_i < K$}
\STATE Sample $(\mathbf{x},\mathbf{y}) \in \mathbf{X}$ s.t. $\mathcal{C}_i \in \mathbf{y}$,  w/o replacement
\STATE $S \gets S \cup \{ (\mathbf{x},\mathbf{y}) \}$
\STATE Update $\{ \text{Count}_j \}$ $\forall \mathcal{C}_j \in \mathbf{y}$
\ENDWHILE
\ENDWHILE
\RETURN $\mathcal{S}$
\end{algorithmic}
\end{algorithm}

\begin{table*}[ht]
{\small
\begin{tabular}{lp{0.9\textwidth}}
\toprule
\standard
&  [SEP] Joe Mack , a district manager for Cormack Enterprises Inc. , a Burger King operator in Omaha , Neb. , says discounting is so prevalent that `` we have to serve 15 \% to 20 \% more customers '' to keep sales level . so prevalent is ADJP . [SEP] As surely as a seesaw tilts , falling interest rates force up the price of previously issued bonds . up is ADVP . [SEP] Pamela Sutherland , executive director of the Illinois Planned Parenthood Council , says she and her allies are `` cautiously optimistic '' they can defeat it if it comes to a floor vote . it is NP . [SEP] An investment group led by Chicago 's Pritzker family recently lowered a \$ 3.35 billion bid for American Medical International , Beverly Hills , Calif. , because of the threat of the legislation . of is PP . [SEP] Joe Mack , a district manager for Cormack Enterprises Inc. , a Burger King operator in Omaha , Neb. , says discounting is so prevalent that `` we have to serve 15 \% to 20 \% more customers '' to keep sales level . that is SBAR . [SEP] Joe Mack , a district manager for Cormack Enterprises Inc. , a Burger King operator in Omaha , Neb. , says discounting is so prevalent that `` we have to serve 15 \% to 20 \% more customers '' to keep sales level . says is VP . \\ \midrule
\standardwrong &  [SEP] Joe Mack , a district manager for Cormack Enterprises Inc. , a Burger King operator in Omaha , Neb. , says discounting is so prevalent that `` we have to serve 15 \% to 20 \% more customers '' to keep sales level . so prevalent is ADVP . [SEP] As surely as a seesaw tilts , falling interest rates force up the price of previously issued bonds . up is NP . [SEP] Pamela Sutherland , executive director of the Illinois Planned Parenthood Council , says she and her allies are `` cautiously optimistic '' they can defeat it if it comes to a floor vote . it is PP . [SEP] An investment group led by Chicago 's Pritzker family recently lowered a \$ 3.35 billion bid for American Medical International , Beverly Hills , Calif. , because of the threat of the legislation . of is SBAR . [SEP] Joe Mack , a district manager for Cormack Enterprises Inc. , a Burger King operator in Omaha , Neb. , says discounting is so prevalent that `` we have to serve 15 \% to 20 \% more customers '' to keep sales level . that is VP . [SEP] Joe Mack , a district manager for Cormack Enterprises Inc. , a Burger King operator in Omaha , Neb. , says discounting is so prevalent that `` we have to serve 15 \% to 20 \% more customers '' to keep sales level . says is ADJP . \\ \midrule
\standardnol &  [SEP] Joe Mack , a district manager for Cormack Enterprises Inc. , a Burger King operator in Omaha , Neb. , says discounting is so prevalent that `` we have to serve 15 \% to 20 \% more customers '' to keep sales level . [SEP] As surely as a seesaw tilts , falling interest rates force up the price of previously issued bonds . [SEP] Pamela Sutherland , executive director of the Illinois Planned Parenthood Council , says she and her allies are `` cautiously optimistic '' they can defeat it if it comes to a floor vote . [SEP] An investment group led by Chicago 's Pritzker family recently lowered a \$ 3.35 billion bid for American Medical International , Beverly Hills , Calif. , because of the threat of the legislation . [SEP] Joe Mack , a district manager for Cormack Enterprises Inc. , a Burger King operator in Omaha , Neb. , says discounting is so prevalent that `` we have to serve 15 \% to 20 \% more customers '' to keep sales level . [SEP] Joe Mack , a district manager for Cormack Enterprises Inc. , a Burger King operator in Omaha , Neb. , says discounting is so prevalent that `` we have to serve 15 \% to 20 \% more customers '' to keep sales level . \\ \midrule
\randomtotally &  [SEP] \#\#llan costs similar Requiem tracking Michelle seeds 15th HM influenced \#\#OH \#\#inia canyon visited USB punished \#\#ter hadn mom \#\#BA \#\#rrow \#\#hetto \#\#loss idea carriages \#\#uk Mellon inconsistent archaeologists Server quartet Low Downs \#\#izations Bears \#\#titis again falsely sprawling Dennis hey plural exam goalkeeper kingdom Argentine befriended \#\#ndi accept 1926 symbolic Colonel [SEP] reviewer sketch rabbi Tampa \#\#orra tour Jul minorities \#\#iary closing Beta Sunday Jai counts quasi \#\#uminous ambitious Funk Got \#\#orm types [SEP] Another Elements growled \#\#aris evaluation resulted announcement Upon complications Brighton \#\#umble \#\#mat compromise grinned Fritz conversations cavalry aids conflicts Kung Bayern Soundtrack \#\#ny remarried \#\#pta indicates cautious \#\#gis arch LP Event clip \#\#ake lobbying Majority Nam select \#\#khar neighbourhood [SEP] Preliminary Barclay \#\#it dialogue html \#\#ce \#\#bul Zimbabwe combines \#\#uch capacity challenged Burgess Stations freestyle vulnerable unbeaten Bordeaux Hyderabad Hearing Zeus Romania \#\#ulate Dead Zhang fullback \#\#kley cruisers Burke Specialist \#\#uy Yuri Trains Cedar Strait rested und Ultra duel attempts Campo [SEP] step landing losses \#\#bones emotions ripe \#\#sad Williamson \#\#MO Tour \#\#DS catalogue \#\#mbs Pietro Text \#\#ão Ada British chalk biologist stating reigned tastes favorites 1839 seduce \#\#track Chilean Arab Johnston \#\#human creative \#\#dox their spotlight subsidies pounding ashore eroded Chart \#\#bull worldwide battled inclusion Verona \#\#bull BP computers losses nurses analogue 1870 [SEP] specialised bony differing I scarce \#\#EN \#\#oring reached peak Lea \#\#stones \#\#press \#\#hall \#\#tic march tuning attacked portion Dietrich Leaving Romani purchased mosques \#\#physical stimulate rejected stages \#\#mart Constance manipulate remote Maya redesigned hazardous seeded usage Scholars follower Endowment Zionist Otto speeding \#\#nch accusations wrath inconsistent Madras Isles classics likewise cousin 360 \\ \midrule
\randomsupport &  [SEP] take \#\#if ' \#\#age \#\#front \% we Mack post Hills firm \#\#age not American \#\#up are P s Medical prevalent to rates preferring ' industry says look takes w not allies that defeat e \#\#ed fur investment said line \#\#ck entrepreneur are 5 \#\#ed \#\#rma prevalent Peter fined Hills Medical director International [SEP] 18 made , preferring sales line family News comes word the executive bid escaped T un From and months wherever \#\#eb [SEP] \#\#o . family cents have Ka Mr bid or to Beverly director \#\#ed if capitalism \#\#sty prison Enterprises Pamela sentenced is dollar level w capital bet 35 talk capitalism Burger The News \% Enterprises floor director Con take an [SEP] entirely that stock false poorly R \#\#nn a have many P line one \#\#rent level 3 so do look one so only falling Pa sound comes district sales Mack 20 prices industry word sales this cloak ` cloak \#\#rent months that [SEP] Mack force \#\#s Medical gone recently \#\#if \#\#rma From \#\#ck \#\#urn capital are \#\#hem 3 Illinois director \#\#hem vote bonds more escaped group capital other because capital word year level R sound 5 \#\#if T optimistic \#\#ck if An other other sales interest Peter operator rates Cal From \#\#har poorly take rates [SEP] talk says \#\#rma Sutherland Peter Enterprises death president bonds An \#\#isms filing tilt by \#\#if months un line pleaded 15 a 15 do bid for \#\#print \#\#aw \#\#rier . said The R threat threat 43 surely be Inc sales bonds talk are so death gone News an discount t guilty \#\#rent dollar \\ \bottomrule
\end{tabular}}
\caption{\textbf{Example demonstrations for different modes constructed with method in Section~\ref{sec:Pathological}}. The dataset used here is CoNLL00 for chunking task. Example demonstrations for \randomtotally\ and \randomsupport\ are shown as a string of tokens.}
\label{tab:demo_example_cnk}
\end{table*}

\end{document}